\documentclass[conference]{IEEEtran}
\IEEEoverridecommandlockouts
\usepackage{cite}
\usepackage{amsmath,amssymb,amsfonts}
\usepackage{algorithmic}
\usepackage{graphicx}
\usepackage{textcomp}
\usepackage{xcolor}
\usepackage{multicol}
\usepackage{booktabs}
\usepackage{pifont}
\usepackage{multirow}
\usepackage{subfigure}
\usepackage{color, colortbl}
\definecolor{Gray}{gray}{0.9}

\newcommand\blfootnote[1]{%
  \begingroup
  \renewcommand\thefootnote{}\footnote{#1}%
  \addtocounter{footnote}{-1}%
  \endgroup
}

\def\BibTeX{{\rm B\kern-.05em{\sc i\kern-.025em b}\kern-.08em
    T\kern-.1667em\lower.7ex\hbox{E}\kern-.125emX}}
\begin{document}

\title{Content-based Graph Privacy Advisor

}

\author{\IEEEauthorblockN{Dimitrios Stoidis}
\IEEEauthorblockA{\textit{Centre for Intelligent Sensing} \\
\textit{Queen Mary University of London}\\
London, UK \\
dimitrios.stoidis@qmul.ac.uk}
\and
\IEEEauthorblockN{Andrea Cavallaro}
\IEEEauthorblockA{\textit{Centre for Intelligent Sensing} \\
\textit{Queen Mary University of London}\\
London, UK \\
a.cavallaro@qmul.ac.uk}
}

\maketitle

\begin{abstract}
People may be unaware of the privacy risks of uploading an image online. In this paper, we present Graph Privacy Advisor, an image privacy classifier that uses scene information and object cardinality as cues to predict whether an image is private.
Graph Privacy Advisor simplifies a state-of-the-art graph model and improves its performance by refining the relevance of the information extracted from the image. We determine the most informative visual features to be used for the privacy classification task and reduce the complexity of the model by replacing high-dimensional image feature vectors with lower-dimensional, more effective features. 
We also address the problem of biased prior information by modelling object co-occurrences instead of the frequency of object occurrences in each class.

\end{abstract}


\section{Introduction}

Sharing images on social media platforms responds to our need for communication and self-expression. However, images may contain personal information that puts at risk the privacy of the photographer and the people involved. As people may be unaware of the privacy risks associated with an image being uploaded online~\cite{DRS, awareness1, awareness2}, it is important to develop predictive models that inform users about private information before sharing. 

Privacy-related information can be extracted from the image itself or from its metadata (e.g.~user-generated tags). 
Zerr et al.~\cite{picalert2012} use SIFT features~\cite{sift} in conjunction with metadata, including the title and user-generated tags.
Dynamic Multi-Modal Fusion for Privacy Prediction (DMFP)~\cite{atongecaragea2019} fuses features from three modalities (object, scene and tag).
A competence estimation of each modality is performed to determine the best modality and fuse its decisions in the last stage to produce a prediction.
DMFP builds upon the Combination model~\cite{atonge_caragea2018} that merges scene-based tags extracted from images with convolutional networks with features of the detected objects in the image. 
The Gated Fusion model~\cite{privacyalert} fuses predictions generated by single-modal models (for objects, scene and tags) and dynamically learns the fusion weights for each modality.  

Some of the above works extract information from user-generated tags~\cite{picalert2012, atonge_caragea2018, atongecaragea2019, privacyalert}.
Instead we focus on content-based image information only, rather than on the associated metadata.
In related work, image features extracted with VGG-16~\cite{vgg-16} are used to extract private information from images~\cite{Xioufis2016PersonalizedPI}.
A combination of hand-crafted and learned features derived from convolutional layers is used in Privacy-Convolutional Neural Network with Hierarchical features (PCNH)~\cite{PCNH}. iPrivacy~\cite{cooccurence} uses multi-task learning to identify sensitive objects in images by considering relationships between objects and their co-occurrence with defined privacy settings.

In this paper, we improve the performance of a state-of-the-art graph-based network~\cite{GIP} by substituting the input information with smaller-size feature vectors obtained by models pre-trained on scene and object detection. We also show that scene information combined with the cardinality of objects (i.e. the number of objects of a specific category) are salient content-based visual features for the image privacy prediction task. Furthermore, we address potentially biased prior information between objects and privacy classes by encoding the co-occurrences of objects. \blfootnote{The code is available at https://github.com/smartcameras/GPA}

\begin{table*}[t]
    \centering
    
    \caption{Privacy datasets and their annotation.}
    \scriptsize
    \begin{tabular}{lcrll}
    \hline
    {\bf Dataset} & {\bf Ref.}& {\bf \# Images} & {\bf Labels} & {\bf Annotation}\\
    \hline
    \multirow{4}{*}{$\star$PicAlert}&   \multirow{4}{*}{\cite{picalert2012}}  &\multirow{4}{*}{37,535}& \multirow{4}{*}{private, public, und.} & ``{\em Private are photos which have to do with the private sphere (like self portraits, family,}\\
    &&&& {\em friends, your home) or contain objects that you would not share with the entire world} \\
    &&&& {\em (like  a private email). The rest are public. In case no decision can be made, the picture }\\
    &&&& {\em should be marked as undecidable}''\\
    
    \hline
    
         \rule{0pt}{2ex}{VISPR}& {~\cite{vispr}}& {22,167} & {private, public} & EU data Privacy~\cite{euprivacy}, US Privacy Act~\cite{usprivacy}, Social network rules~\cite{socialrule1}\\
        
         \hline
        \rule{0pt}{2ex}{$\star$IPD} &  {~\cite{GIP}}  &{38,525}& private, public, und. & {see PicAlert and VISPR}\\
     
         \hline
         \multirow{3}{*}{$\star$PrivacyAlert} & \multirow{3}{*}{\cite{privacyalert}}  & \multirow{3}{*}{6,800} & {$\dagger$clearly private,} & ``{\em Assume you have taken these photos, and you are about to upload them on your}   \\ 
         && &  clearly public, & {\em favourite social network} [...] {\em tell us whether these images are either private or }  \\
         &&& private, public & {\em public in nature. Assume that the people in the photos are those that you know} '' \\
         \hline
         \multicolumn{5}{l}{{\rule{0pt}{2.2ex}\scriptsize{KEY -- und.: undecidable. VISPR: Visual Privacy dataset~\cite{vispr}, IPD: Image Privacy Dataset~\cite{GIP}, $\star$: full dataset not available,}}}\\
         \multicolumn{5}{l}{{\scriptsize{$\dagger$: classes are merged into two classes (private and public).}}}
    \end{tabular}
    
    \label{tab:sotadatasets}
\end{table*}

\section{Related Work}

In this section, we discuss image privacy datasets and graph-based learning methods for image classification. 

\subsection{Image privacy datasets and annotation}
\label{sec:data}

There exist four main image privacy datasets, namely PicAlert~\cite{picalert2012}, Visual Privacy (VISPR)~\cite{vispr}, Image Privacy Dataset (IPD)~\cite{GIP} and PrivacyAlert~\cite{privacyalert}. {PicAlert}~\cite{picalert2012} has 37,535 images that were posted on Flickr from January to April 2010\footnote{Note that 9,365 images have been deleted after the original publication due to copyright issues.}. In PicAlert, most private images contain people, which induces a bias towards the considered private objects (63\% of private images contain persons). VISPR~\cite{vispr} (22,167 images from Flickr) includes images from natural everyday scenes, with background and foreground clutter, and images with textual content such as ID cards, bank account details, email content, and licence plates. IPD~\cite{GIP} (38,525 images) combines PicAlert~\cite{picalert2012} and 6,392 private images from VISPR~\cite{vispr} (13,910 private images and 24,615 public images). {PrivacyAlert}~\cite{privacyalert} contains 6,800 images that were posted between 2011 and 2021 (83\% dating after 2015, thus more representative of recent trends in image sharing).

Annotations for privacy datasets are notoriously difficult to obtain due to the inherent subjectivity of this labelling process. For PicAlert, participants were asked to judge whether an image was considered in the private sphere of the photographer (e.g. selfies, family, friends and home) (see Tab.~\ref{tab:sotadatasets}). Possible labels are \emph{private}, \emph{public} and \emph{undecidable} (if no decision could be made on the image label). For VISPR, the annotations are based on 68 attributes compiled from the EU Data Protection Directive 95/46/EC~\cite{euprivacy}, the US Privacy Act 1974~\cite{usprivacy}, social network platform rules~\cite{socialrule1} and additional attributes after manual inspection of the images~\cite{vispr}. Private images contain at least one out of 32 privacy attributes related to personal life, health, documents, visited locations, Internet conversations, and automobiles.
For {PrivacyAlert}, the images are annotated through the Mechanical Turk\footnote{https://www.mturk.com/} crowd-sourcing platform, and the annotators are asked to classify the images into 4 classes (\emph{clearly private}, \emph{private}, \emph{public} and \emph{clearly public}). The quality of the annotations is monitored using an attention checker set that discards annotators who failed to provide the expected response. The four-class annotations are then grouped into binary labels that combine \emph{clearly private} with \emph{private} and \emph{clearly public} with \emph{public}. 
{PrivacyAlert} provides binary labels for each image and VISPR provides privacy attributes that are classified as private or public.
In PicAlert, 17\% of the dataset contains multiple ternary annotations for each image where annotators' agreement needs to be computed. Thus, labels can be decided depending on the desired level of privacy, for instance, labelling an image as private in case of annotation disagreement.
%

\subsection{Graph-based models for image classification}

Graph-based methods have recently been introduced for privacy classification~\cite{GIP}, \cite{drag}.
Graph-based networks model information as nodes whose relationship is defined through edges.
The representation of each node is updated, propagating the information through the edges.
The initialisation of graphs is often referred to as prior knowledge represented as an adjacency matrix.

Prior knowledge structured as a knowledge graph can improve image classification performance~\cite{GSNN}.
The Graph Search Neural Network (GSNN)~\cite{GSNN} incorporates prior knowledge into Graph Neural Networks (GNN)~\cite{gnn} to solve a multi-task vision classification problem (see Tab.~\ref{tab:gnn_arch}). GSNN is based on the Gated Graph Neural Network (GGNN)~\cite{ggnn}, reducing computational requirements and observing the flow of information through the propagation model. GGNN uses the Gated Recurrent Unit (GRU)~\cite{gru} to update the hidden state of each node with information from the neighbouring nodes.
GGNN is a differential recurrent neural network that operates on graph data representations, iteratively propagating the relationships to learn node-level and graph-level representations.
The Graph Reasoning Model (GRM)~\cite{GRM} uses objects and interactions between people to predict social relationships.
The graph model weighs the predicted relationships with a graph attention mechanism based on Graph Attention Networks (GAT)~\cite{gat}. 
GRM uses prior knowledge on social relationships, co-occurrences and objects in the scene as a structured graph.
Interactions between people of interest and contextual objects are modelled by GGNN~\cite{ggnn}, where nodes are initialised with the corresponding semantic regions.
The model learns about relevant objects that carry relevant task-information. 

Graph Image Privacy (GIP)~\cite{GIP} replaces the social relationship nodes with two nodes representing the privacy classes (private, public). 
Dynamic Region-Aware Graph Convolutional Network (DRAG)~\cite{drag} adaptively models the correlation between important regions of the image (including objects) using a self-attention mechanism. DRAG~\cite{drag} is a Graph Convolutional Network (GCN)~\cite{gcn} that learns the relationships among specific regions in the image without the use of object recognition.

\begin{figure}[t]
    \centering
    \subfigure[]{\includegraphics[width=4cm,height=3cm]{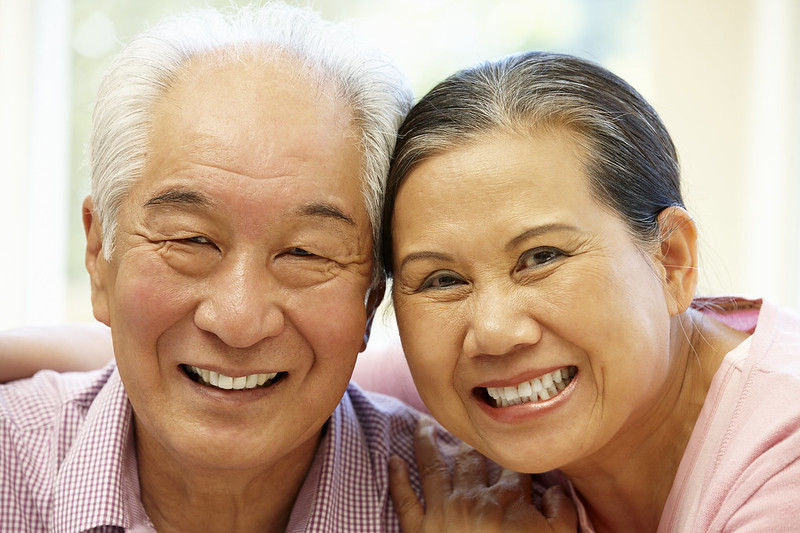}} 
    \subfigure[]{\includegraphics[width=4cm,height=3cm]{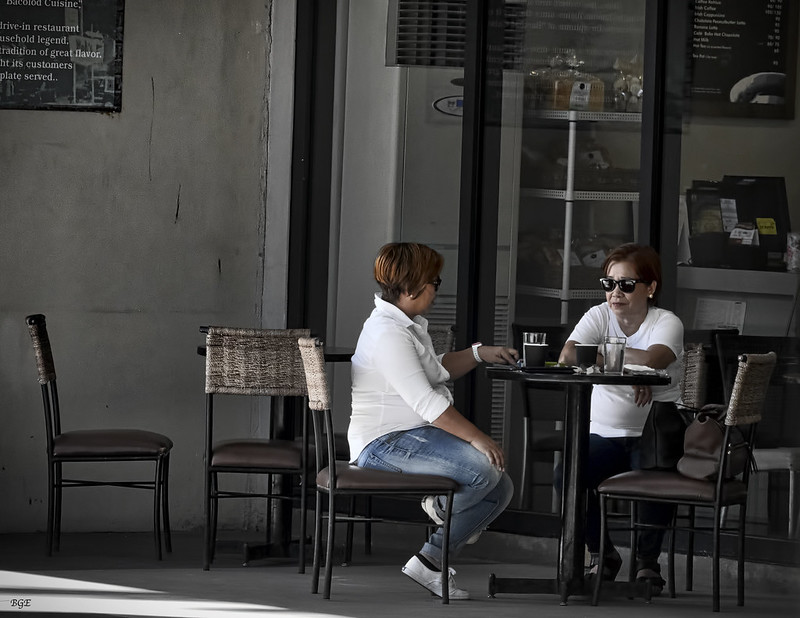}} 
    \subfigure[]{\includegraphics[width=4cm,height=3cm]{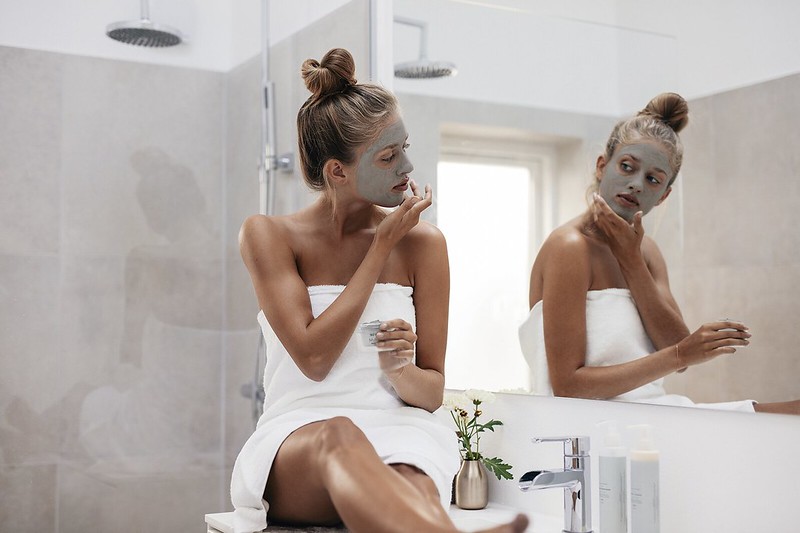}} 
    \subfigure[]{\includegraphics[width=4cm,height=3cm]{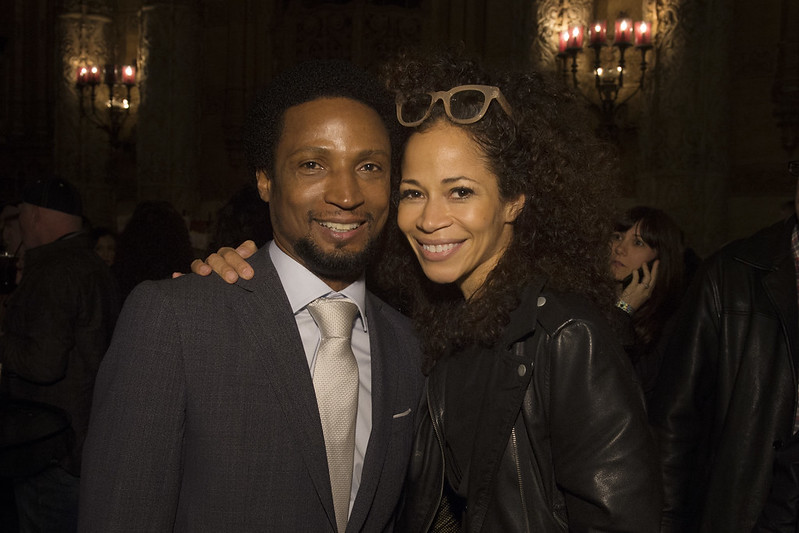}} 

    \caption{Images might be labelled by human annotators as private or public depending on the scene and objects depicted. Images (a) and (c) were labelled as private while (b) and (d) were labelled as public (PrivacyAlert~\cite{privacyalert}).}
    \label{fig:examples}
\end{figure}

\begin{table}[t]
    \centering
        \caption{Main components of the architecture of graph-based methods.}.
    \begin{tabular}{lcll}
    \hline
         {\bf Model} & Ref. & {\bf Architecture} & {\bf Task}\\
         \hline
         GGNN & \cite{ggnn} & GRU+GNN& representation learning\\
         GSNN & \cite{GSNN} & GGNN & image classification\\
         GRM & \cite{GRM} & GGNN+GAT & relationship recognition\\
         GIP & \cite{GIP} & GGNN+GAT & image privacy classification\\ 
         DRAG & \cite{drag} & GCN & image privacy classification \\
         \hline
         
         \multicolumn{4}{l}{\parbox{0.95\columnwidth}{\rule{0pt}{2.2ex}\scriptsize{KEY -- GRU: Gated Recurrent Unit~\cite{gru}; GNN: Graph Neural Network~\cite{gnn}; GAT: Graph Attention Networks~\cite{gat}; GCN: Graph Convolutional Network~\cite{gcn}.}}}\\
    \end{tabular}
    \vspace{-10pt}
    \label{tab:gnn_arch}
\end{table}

\section{Method}
\label{method}

In this section, we present the content-based features, the prior information to initialise Graph Privacy Advisor (GPA) as well as the graph-based learning and privacy classification.

\begin{figure*}\label{fig:arch}
    \centering
    \includegraphics[width=\linewidth]{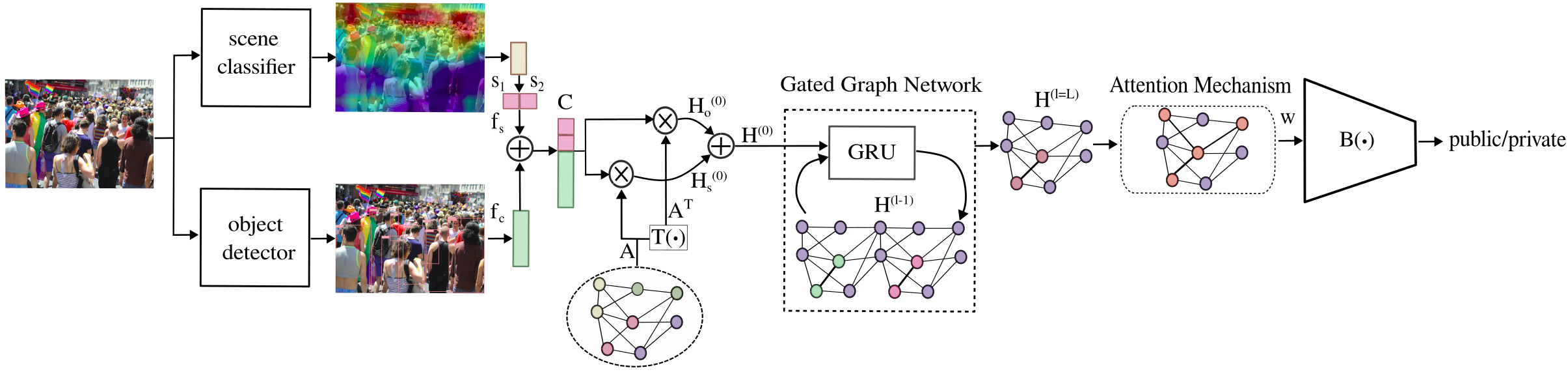}

    \caption{The Graph Privacy Advisor model. Scene information $f_s$ is extracted from the input image with a scene classifier. The co-occurrence and the cardinality of objects in each image $f_{c}$ are obtained using an object detector. Scene information concatenated with object cardinality is combined with the prior knowledge (A) and its transpose $A^T$, producing two graphs $H^{(0)}_o$ and $H^{(0)}_s$ that are concatenated into the hidden state matrix $H^{(0)}$ and provided to the Gated Graph Network. After $L$ iterations, the learned representation of nodes $H^{(l=L)}$ undergoes an attention mechanism that weighs the most relevant object nodes. The final binary classification $B(\cdot)$ produces the prediction scores for each privacy class. KEY-- GRU: Gated Recurrent Unit, $\otimes$: matrix multiplication, $\oplus$: concatenation.}
\end{figure*}

\subsection{Features}

Cardinality can affect the prediction of the privacy class, especially considering the \emph{person} category. An image is more likely to be public if the cardinality of \emph{person} is high. For instance, in Fig.~\ref{fig:examples} images (a) and (d) both depict a couple posing indoors, however (a) is private, whereas (d) is public. In settings where the same objects and the same cardinality are depicted, the context of the scene might provide additional information for the task.
We explicitly incorporate object cardinality information using a pre-trained object detection model.
We define the cardinality vector, $f_{c} \in \mathbb{R}^K$, where $K$ is the number of object categories. 

We extract scene information with a scene recognition model that we train to differentiate between two privacy classes instead of 365 scene classes~\cite{sceneplaces}. A fully connected layer introduced at the output of the scene recognition model takes the logits produced by the model and reduces the dimensionality to two logit values, which we denote as $s_{1}$ and $s_{2}$. We define the scene vector $f_s \in \mathbb{R}^2$ as $f_s=\big(s_1, s_2\big)$.

We define a matrix incorporating scene and object information for each image.
Let $C \in \mathbb{R}^{2 \times (2+K)}$ be the information matrix defined as:
\begin{equation}
  C = \big(f_h \oplus (f_s \oplus f_{c})\big),  
\end{equation}
where $f_h$ is a one-hot vector that differentiates between scene and object information and $\oplus$ is the concatenation operation. The size of $C$ is $2 \times (2+K)$, where 2 is the number of privacy classes and $K$ the number of object categories.

The matrix $C$ greatly simplifies the information used in GIP~\cite{GIP}.
We replace the high-dimensional feature vectors obtained from feature extracted from pre-trained models with a single cardinality vector and two scene logits. This replaces the 4096-dimensional feature vectors with a single value. These high-dimensional vectors are used by Yang et al.~\cite{GIP} to encode two types of information:
one from the entire image and the other from the individual detected objects. Section~\ref{val} shows that the information used by Yang et al.~\cite{GIP} does not improve the performance of the model and greatly increases the model size.

\subsection{Prior knowledge}
\label{priorknowledge}

Prior information can be defined by the initial edges that convey information between the input nodes and is encoded in an adjacency matrix $A$.

We aim to model relationships between objects co-occurring in images to uncover privacy-exposing patterns. 
We hypothesise that considering prior knowledge modelling relationships between object nodes and privacy nodes biases the model towards more frequent objects found in the data.
To avoid this problem, we therefore propose using a binary matrix of object co-occurrences to minimise the effect of frequent objects, such as \emph{person}, being associated with the private class.

Let us consider $K$ object categories and let 
\begin{equation}
a_{ij}=
\left\{
    \begin{array}{ll}
         1 & \text{if object $i$ and object $j$ co-occur} \\
         0 & \text{otherwise}\\
    \end{array}
\right .
\end{equation}
be elements of a square binary matrix $A^{K}=(a_{ij})_{K\times K}$ that defines the co-occurrence of object categories $i$ and $j$ in an image. 
As the model learns to discriminate between private and public images during training, the object co-occurrences are associated with the privacy classes based on their frequency of appearance in each image. Therefore, the presence of an object pair will affect the prediction of the privacy class.
Considering the co-occurrence of objects gives more importance to the presence of less frequent objects such as a \emph{tie} which is more likely to be associated with public events (see Fig.~\ref{fig:examples} (a) and (d)).

Let $A^{K+2}$ be the matrix $A^{K}$ augmented with two rows and two columns of zeros.
We pad the matrix to account for information on the scene representing the two privacy classes to be encoded before learning the graph. This information is provided by combining prior knowledge with information for each image during training. The simplified notation $A$ is used to refer to matrix $A^{K+2}$ in the rest of the paper.
Unlike in GIP~\cite{GIP}, in GPA, the matrix $A$ does not encode prior information on the privacy class of $K$ object categories. 
The initialisation of GPA addresses biased information relating to the long-tail distribution~\cite{long-tail} of object categories found in the dataset by considering object co-occurrences instead of the frequency of occurrence (recurrence) of objects.

\subsection{Graph learning} 

Information is propagated in the graph through message passing that updates the state of nodes aggregating information from neighbouring nodes.
The matrix $C$, which encodes the scene and cardinality information for each image, is multiplied by the prior information, modelled by the adjacency matrix $A$ and its transpose $A^T$.
The initial state of the graph at $l=0$ with respect to object cardinality can be represented as 
\begin{equation}
    H_o^{(0)} = A \otimes C,
\end{equation}
whereas the initialisation with scene information gives
\begin{equation}
    H_s^{(0)} = A^T\otimes C,    
\end{equation}
 where $\otimes$ denotes matrix multiplication.
$H_o^{(0)}$ and $H_s^{(0)}$ both in $\mathbb{R}^{2\times (K+2)}$ encode the initial state of the nodes.
The hidden state matrix that performs the message passing operation is modelled by the matrix $H^{(l)}$ for each iteration $l$.
In the initial state $l=0$, $H^{(0)}$ concatenates the result of these two operations:

\begin{equation}
    H^{(0)} = H_o^{(0)} \oplus H_s^{(0)},
\end{equation}
where $\oplus$ is the concatenation operation and $H^{(0)} \in \mathbb{R}^{4 \times (K+2)}$.

For learning, GPA maintains components of GIP~\cite{GIP}, namely, GGNN~\cite{ggnn} and the GAT-based attention mechanism~\cite{gat} (see Fig. 2).
GGNN~\cite{ggnn} is used to iteratively learn the relationships between the nodes, given prior knowledge.
The hidden state matrix $H^{(l)}$ is updated through a GRU layer~\cite{gru} where messages from neighbouring nodes and previous iterations are aggregated to update the hidden state of each node for each iteration $l$. 
The learned node states and edges at the output of the gated network are subsequently weighted with a graph attention mechanism that determines the most relevant output nodes based on their importance.
The edges between object node $v_i$ and object node $v_j$ are weighted with attention coefficients.
The coefficients are obtained by applying the attention mechanism $\alpha(\cdot)$ on the object nodes $e_{ij} = \alpha(v_{i},v_{j})$ and are normalised using a sigmoid function. 
The resulting weighted feature vector $w$ is first reshaped with a linear layer and then fed through a classifier $B(\cdot)$ that yields a prediction score for each class. 

The classifier $B(\cdot)$ consists of two fully connected layers $g_1(\cdot)$ and $g_2(\cdot)$, followed by a {\tt softmax} operation:
\begin{equation}
   B(w) = {\text{\tt softmax}}\Big(g_{2}\big(g_1(w)\big)\Big).
\end{equation}
To learn the binary classification task, we use the cross entropy as the loss function:
\begin{equation}
    \mathcal{L} = -\sum_{n=1}^N [y_n \log(p_n) + (1-y_n)\log(1-p_n)],
\end{equation}
where $y_n$ is the ground-truth class label for image $n$, $p_n$ is the corresponding model prediction, and $N$ is the total number of images in the set.

\section{Validation}\label{val}

In this section, we analyse the impact of different graph initialisations and considered features.
We compare the proposed model, GPA, with content-based vision models, namely PCNH~\cite{PCNH}, GIP~\cite{GIP}, and DRAG~\cite{drag}. 

\subsection{Dataset and annotation}

We use three datasets, namely PicAlert2, which consists of 28,170 images (20,555 public and 7,615 private images); IPD2, which comprises 20,555 public and 14,007 private images; and PrivacyAlert2, which has 6,476 images (4,959 public and 1,517 private images). 
IPD2 includes public images from PicAlert2 and 6,392 private images of VISPR.
Due to the reasons discussed in Section~\ref{sec:data}, PicAlert2 and IPD2 contain 4,000 fewer images with respect to the datasets used by Yang et al.~\cite{GIP}, while PrivacyAlert2 contains 357 fewer images than in the dataset used by Zhao et al.~\cite{privacyalert}.

Because of the limitations in the annotation discussed in Section~\ref{sec:data}, for PicAlert2 we consider images that have unambiguous annotations (83\% of the dataset) and label the remaining images as public when the ratio of public votes over private votes is close to 3. The remaining images were considered private.
We obtain a dataset with a ratio of public to private images of 2.7:1 in contrast with 3.3:1 in Yang et al.~\cite{GIP}. 
Thus we include more images as private by considering images with doubtful annotations as private, favouring sensitivity.

\begin{table*}[t]
    \centering
    
        \caption{Impact of different types of prior knowledge for the GPA model tested on the PrivacyAlert2 dataset.}

    \begin{tabular}{ccccccccccc}
    \toprule
         &  \multicolumn{3}{c}{Public}  & & \multicolumn{3}{c}{Private} & & \multicolumn{2}{c}{Overall} \\
         \cmidrule[0.5pt]{2-4}\cmidrule[0.5pt]{6-8} \cmidrule[0.5pt]{10-11}
         Adjacency matrix  & P & R & 
        F1 && P & R & F1 && UBA(\%) & U-F1  \\
        \midrule
         Random generator & .770 & .520 & .621 && .255 & .514 & .341 && 51.90 & .481 \\
         \midrule

         ${{\tt uniform}}$  &  {.789} & {.995} & .880 && {.910} & {.169} & .285 && 79.47 &.582  \\
         ${{\tt random}}$ &  {.845} & .907 & .875 & & .623& {.481} & {.543} && 80.39 & .709 \\

        ${\tt ones}$ &  .859 & .903 & .880 && .640 & .538 & .585 && 81.48 & .732\\
         
        ${\tt class}$ &  {.856} & .935 & \textbf{.894} &&  .715 & {.507} & .593 && 83.16 & .744  \\
         ${\tt co\textrm{-}occurrence}$ &  {.857} & .935 & \textbf{.894} && .715 & .514 & \textbf{.598} && \textbf{83.28} & \textbf{.746}\\
        \hline
        \multicolumn{11}{l}{\rule{0pt}{2.2ex}\scriptsize{UBA(\%): unweighted binary accuracy, P: Precision, R: Recall, U-F1: unweighted F1 score.}}\\
        \multicolumn{11}{l}{\scriptsize{Random generator: baseline test with a random number generator.}}\\
        \multicolumn{11}{l}{\scriptsize{${\tt uniform}$: uniform value assuming equi-probable frequency of occurrence for each object in each privacy class.}}\\
        \multicolumn{11}{l}{\scriptsize{${\tt ones}$: matrix filled with ones.}}\\
        \multicolumn{11}{l}{\scriptsize{${\tt random}$: matrix with random values.}}\\
        \multicolumn{11}{l}{\scriptsize{${\tt class}$: matrix of object recurrences in each privacy class.}}\\
        \multicolumn{11}{l}{\scriptsize{${\tt co\textrm{-}occurrence}$: binary matrix of object co-occurrences.}}\\

    \end{tabular}
    \label{tab:results_adj_init}
\end{table*}

\begin{table*}[t]
    \centering
        \caption{Impact of different features when training GPA on PrivacyAlert2. Initialisation was performed with object co-occurrences.}
    \begin{tabular}{cccccccccccccc}
    \toprule
         &  & & & \multicolumn{3}{c}{Public}  & & \multicolumn{3}{c}{Private} & & \multicolumn{2}{c}{Overall} \\
        \cmidrule[0.5pt]{5-7}\cmidrule[0.5pt]{9-11} \cmidrule[0.5pt]{13-14}
        $f_i$ & $f_s$ & $f_o$ & $f_c$ &  P & R & F1 & & P & R & F1 && UBA(\%) & U-F1\\
        \midrule
         $\circ$ & $\circ$ & $\circ$ & $\circ$ & .758 & {1.00} & .862 && \textendash & .000 &  .000 && 75.78 & .431\\
         
         $\bullet$ & $\circ$ & $\circ$ & $\circ$  &.758 & {1.00} & .862 & & \textendash & .000& .000 &&  75.78 & .431\\
        \rowcolor{Gray}
         $\circ$ & $\circ$ & $\circ$ & $\bullet$&  .815 & .964 & .883 & & .737 & .314 & .440 && 80.68 &.661 \\
        
        $\circ$&  $\circ$& $\bullet$ & $\bullet$ &   .830 & .909 & .868 && .595 & .417 & .490 && 79.01 &.679\\
        
        $\bullet$ & $\circ$ & $\bullet$ & $\circ$ &  {.924} & .687 & .788 &  & .457 & {.824} & .588 && 72.03 &.688\\
        
        $\circ$ & $\circ$ & pers. & $\circ$ &  .851 & .877 & .864  && .575 & .519 & .512 && 79.06 &.688\\
         
        $\bullet$ & $\bullet$ & $\circ$ & $\bullet$ &  .849 & .868 & .858 && .555 & .517 & .535 && 78.26 &.697\\
        
        $\circ$ & $\bullet$& $\circ$ & $\circ$ &  .838 & .937 &.885 && .687 & .433 & .531 && 81.49 &.708\\
        
        $\bullet$ & $\circ$ & $\bullet$ & $\bullet$ &  .866 & .852 & .859 & & .558 & .588 & .573 && 78.77 & .716\\
        
        $\circ$ & $\circ$ & $\bullet$& $\circ$ &  .896 & .802 & .846 & & .534 & .709 & .609 && 77.97 &.727 \\
        \rowcolor{Gray}
        $\circ$ &  $\bullet$ & $\circ$ & $\bullet$ &  {.857} & {.935} & \textbf{.894} && {.715} & {.514} & {.598} && \textbf{83.28} & .746\\
        
        $\bullet$ & $\bullet$ & $\bullet$ & $\bullet$ &  .891 & .833 & .861 & & .565 & .681 & .617 && 79.58 & .771\\
        
        $\circ$ & $\bullet$ & $\bullet$ & $\bullet$ &  .909 & .814 & .859 && .562 & .745 & \textbf{.642} && 79.76 & \textbf{.802}\\
        
        \hline
        \multicolumn{14}{l}{\rule{0pt}{2.2ex}\scriptsize{$f_i$ : whole-image features, $f_s$: scene information, $f_o$: object features, $f_c$: object cardinality}}\\
        \multicolumn{14}{l}{\scriptsize{UBA(\%): unweighted binary accuracy. P: Precision, R: Recall, U-F1: unweighted F1 score.}}\\ 
        \multicolumn{14}{l}{\scriptsize{pers.: features of the category of objects \emph{person}}.}
        
    \end{tabular}
    \label{tab:object_feat}
\end{table*}

\subsection{Performance measures and comparison} \label{perfmeas} 

We evaluate the results in terms of unweighted (overall) binary accuracy (UBA), precision, recall, F1 score for the private and public class, and unweighted F1 score (U-F1). 

UBA is the ratio between the correctly classified images of both classes and all the images in the test set.
The precision of the private (public) class is the ratio between the number of images correctly classified as private (public) over those predicted as private (public). Recall is the ratio between correctly classified private (public) images and all private (public) images. F1 score is the harmonic mean of precision and recall. U-F1 is the mean of the F1 scores in each class.
F1 score is a useful measure to evaluate performance in the binary classification of imbalanced data.
Per-class measures are more important than overall results, as a lower false negative rate is desired such that fewer private images are predicted as public.
To minimise the risk of personal information leakage, we consider the {per-class F1 scores and unweighted F1 as the most important measures}.

We compare our results with vision-based models on the binary image privacy classification task.
The performance values for the PCNH~\cite{PCNH} and GIP~\cite{GIP} models in PicAlert and the Image Privacy Dataset were taken from Yang et al.~\cite{drag}. The performance results for PCNH~\cite{PCNH} (re-trained and evaluated) in the PrivacyAlert dataset were taken from Zhao et al.~\cite{privacyalert}.

We train and test the GIP model following the partitions of train, validation and test data in~\cite{GIP, drag}. We use random sampling, preserving the ratio of images in the train and the validation sets.
We perform 3-fold cross-validation and train for a total of 50 epochs using a batch size of 64. The categorical cross-entropy loss is optimised with the Adam~\cite{adam} optimiser.
We use PyTorch~\cite{pytorch} 1.8.1 and a single 32GB Tesla V100 Graphical Processing Unit (GPU) with an initial learning rate of 0.0001.

We also consider a random number generator to determine baseline classification results for the imbalanced classes in the dataset.

\begin{table*}[t]
    \centering
        \caption{Performance of image-based privacy classification methods. Results for related work are taken from the literature~\cite{drag, privacyalert}. }
    \begin{tabular}{llccccccccccc}
    \toprule

         &   &   \multicolumn{3}{c}{Public} & &  \multicolumn{3}{c}{Private} && \multicolumn{2}{c}{Overall} \\ \cmidrule[0.5pt]{3-5}\cmidrule[0.5pt]{7-9} \cmidrule[0.5pt]{11-12}
        
        Dataset & Model &  P & R & F1 & & P & R & F1 && UBA(\%) & U-F1 \\
        
         \midrule
        \multirow{3}{*}{PicAlert} & PCNH~\cite{PCNH}  &  .862 & {.929 }& .894 & & .689 & .514 & .589 && 83.15 & .741 \\
        & DRAG~\cite{drag} &  .914 & .914 & \textbf{.914} && {.719} & {.719} & \textbf{.719} && \textbf{86.84} & \textbf{.816} \\
        & GIP~\cite{GIP} &  {.922} & .871 & .895 & & .522 & .684 & .610 && 83.49 & .753 \\
        \cmidrule[0.3pt]{1-12}
        \multirow{3}{*}{PicAlert2} & Random gen. & .748 & .489 & .591 && .248 & .505 & .333 && 49.33 &  .462 \\
        & GIP$^{tr}$~\cite{GIP}  &  .872 & .882 & .877 && .670 & .649 & .659 && {81.89} & .768\\
        & GPA (ours) &  {.869} & {.917} & \textbf{.892} && {.737} & {.626} & \textbf{.677} && \textbf{83.87} & \textbf{.784}\\
        
        \midrule[1.0pt]
        
        \multirow{1}{*}{PrivacyAlert} & PCNH~\cite{PCNH} &  .851 & .929 & .888 && .706 & .511 & .593 && {83.17} & .740 \\

        \cmidrule[0.3pt]{1-12}
        
        \multirow{4}{*}{PrivacyAlert2} & Random gen. &  .770 & .520 & .621 && .255 & .514 & .341 && 51.90 & .481 \\
        & GIP$^{tr}$~\cite{GIP} &  .838 & {.902} & .869 && .597 & .455 & .516 && 79.35 & .693\\
         & GPA (ours) &  {.857} & {.935} & \textbf{.894} && {.715} & {.514} & \textbf{.598} && \textbf{83.28} & \textbf{.746}\\
        \midrule[1.0pt]
        \multirow{2}{*}{Image Privacy Dataset} & GIP~\cite{GIP}  &  .730 & .795 & .761 & & {.812} & .751 & {.780} && 77.09 & .770\\
        & DRAG~\cite{drag} & {.914} & {.895} & \textbf{.905} && .811 & {.842} & \textbf{.826} && \textbf{ 87.68} & \textbf{.865}\\
        \cmidrule[0.3pt]{1-12}
        \multirow{4}{*}{Image Privacy Dataset2} & Random gen.  &   .686 & .494 & .574 && .316 & .508 & .389 && 49.87 & .481\\
        & GIP$^{tr}$~\cite{GIP} &  {.856} & {.901} & \textbf{.878} & & .689 & .590 & .636 && \textbf{81.72} & .757 \\
        & GPA (ours) &  .827 &{.860} & {.843} && {.782} & {.736} &\textbf{.758} && {80.96} & \textbf{.800} \\
        \hline
        \multicolumn{11}{l}{\rule{0pt}{2.2ex}\scriptsize{UBA(\%): unweighted binary accuracy. P: Precision, R: Recall, U-F1: unweighted F1 score.}}\\
        \multicolumn{11}{l}{\scriptsize{Random gen.: random generator, $^{tr}$: trained of the $\star$2 dataset.}}\\
    \end{tabular}
    \label{tab:SOTA_IPD}
\end{table*}

\subsection{Discussion}

Table~\ref{tab:results_adj_init} shows the impact of graph initialisation using different adjacency matrices, namely $\tt uniform$, $\tt random$, $\tt ones$, $\tt class$ and $\tt co\textrm{-}occurrence$.

A uniform initialisation considers objects as equi-probable ($\frac{1}{K}$) to occur in each class. The total number of categories of objects considered is $K=81$~\cite{GIP}. Although performance in the public class is comparable to other methods, results in the private class show a low recall value (0.169), denoting a large number of miss-classifications of private images as public. Initialising $A$ with ${\tt random}$ values or ${\tt ones}$ (matrix filled with ones) yields comparable performance, with both initialisations scoring higher in the public class than in the private one. Assuming only binary information on object co-occurrences, with the ${\tt co\textrm{-}occurrence}$ matrix, shows the best results both in terms of overall performance (UBA of 83.28 and U-F1 of 0.746) and per-class F1 scores (0.894 for public and 0.598 for private). The $\tt class$ adjacency matrix used in GIP~\cite{GIP} shows comparable performance with the ${\tt co\textrm{-}occurrence}$ matrix used in GPA in the public class (both have F1 score of 0.894). However, performance in the private class is slightly improved by using the ${\tt co\textrm{-}occurrence}$ matrix as seen by the F1 value (0.598 versus 0.593). Note that the effect of $\tt random$ initialisation is the most significant as the impact on overall performance is limited.

Table~\ref{tab:object_feat} compares the impact of the different features: features from the whole image ($f_i$), features of the detected objects ($f_o$), the information of the scene ($f_s$) and the cardinality of the objects in a category ($f_c$).

We use a ResNet-101~\cite{resnet101}, pre-trained on ImageNet~\cite{imagenet} to extract the high-dimensional features of the entire image $f_i$.
The features $f_o$, encoding the detected objects in each image using the VGG-16~\cite{vgg-16} model, pre-trained on ImageNet~\cite{imagenet} for the object detection task. VGG-16 extracts the feature vectors of objects using information on the Regions of Interest (ROIs) in the image, detected using the YOLOv3 object detection model~\cite{yolov3}.
We set a confidence threshold for ROI selection to 0.8 and consider a maximum of 12 objects per image, as in Yang et al.~\cite{GIP}.
Each ROI is labelled with 80 object categories from the COCO~\cite{coco} dataset with the addition of the \emph{background} category to account for images with no apparent objects.
Both $f_o$ and $f_i$ are $d$-dimensional features, where $d=4096$. Therefore, $C\in \mathbb{R}^{d\times (2+K)}$ in GIP~\cite{GIP}.

The vector $f_c$, which encodes the cardinality of objects in each category, is obtained from the pre-computed object information (category and number) from YOLOv3 without the need to perform object detection online.
To incorporate scene information we use a pre-trained ResNet-50, referred to as Places365, on scene recognition, that classifies images in 365 pre-defined scene categories~\cite{sceneplaces}. We add a fully-connected layer before the final softmax layer of the ResNet-50 model to compress the output into two categories and use the output logits as scene information.

Table~\ref{tab:object_feat} shows that using the whole-image features $f_i$ does not improve performance.
In the first two cases (with and without $f_i$), F1 score for the private class is zero, predicting all instances as public. 
$f_o$ and $f_i$ used together (in GIP) lead to the lowest UBA (72.03\%) and lowest F1 score in the public class (0.788). 
A \emph{person} detector encoding exclusively the presence of the \emph{person} object (denoted as `pers.' in Tab.~\ref{tab:object_feat}) produces better results than considering all possible objects $f_o$ with cardinality $f_c$ and $f_o$ with whole-image features $f_i$.
The combination of all four features produces the second-best overall performance in terms of U-F1 (0.771), while removing $f_i$ and keeping the remaining features, presents the best performance in the private class ($F1 = 0.642$) and the best overall performance with $U\textrm{-}F1 = 0.802$.
Although cardinality ($f_c$) and scene information ($f_s$) perform poorly overall when used alone, their combination yields the best overall performance in UBA (83.28) and the third best in U-F1 (0.746).
Therefore, we consider $f_c$ and $f_s$ for the privacy classification task, as they also have a smaller dimensionality compared to $f_o$ and $f_i$.

In Table~\ref{tab:SOTA_IPD}, GPA outperforms the trained GIP$^{tr}$~\cite{GIP} model in all the datasets. In PicAlert2 and PrivacyAlert2, GPA outperforms GIP$^{tr}$ both in terms of F1 scores in privacy classes and overall in terms of UBA and U-F1 score. Specifically, for PrivacyAlert2, GPA greatly improves overall performance against GIP$^{tr}$ with a UBA value of 83.28 (79.35 in GIP$^{tr}$) and U-F1 of 0.746 (0.693 in GIP$^{tr}$).
However, performance in the private class remains low, with a recall value of 0.514, the same as that of the random generator.
In IPD2, GPA scores lower in the public class compared to GIP$^{tr}$ (F1 score of 0.843 versus 0.878), however significantly improves performance in the private class (F1 score of 0.758 versus 0.636). In terms of overall performance, GIP$^{tr}$ scores higher than GPA in UBA (81.72 against 80.96), though we favour the U-F1 score over the UBA measure (see Section~\ref{perfmeas}).

\begin{figure}[t]
    \centering

    \subfigure[]{\includegraphics[width=2.7cm,height=3.5cm]{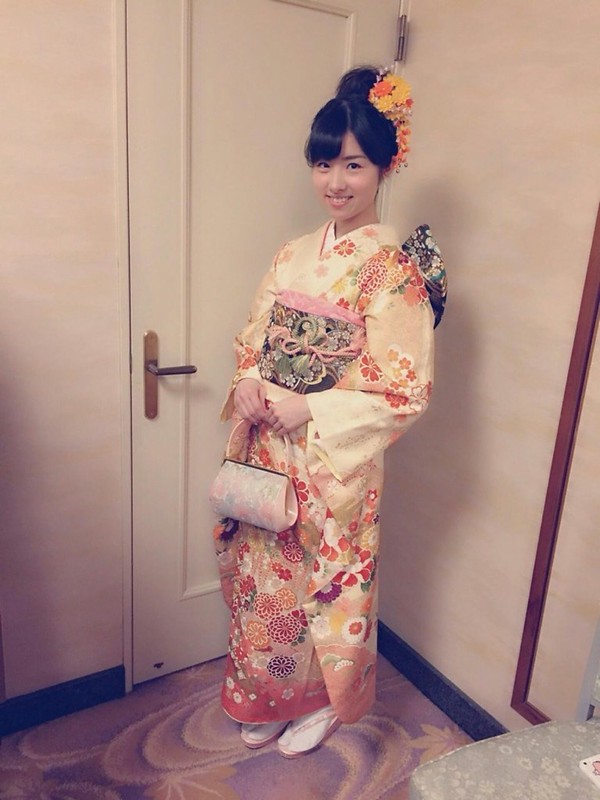}} 
    \subfigure[]{\includegraphics[width=2.7cm,height=3.5cm]{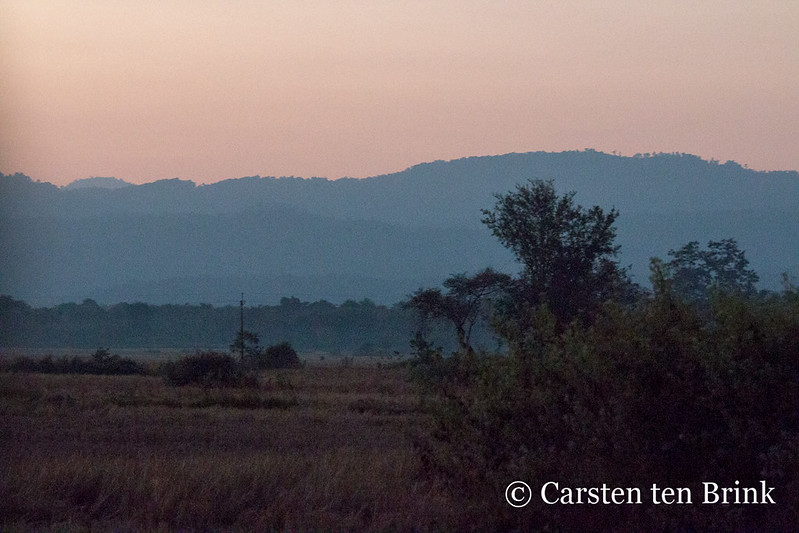}} 
    \subfigure[]{\includegraphics[width=2.7cm,height=3.5cm]{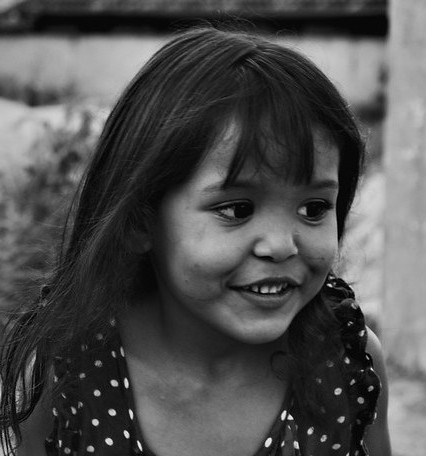}} 
    \subfigure[]{\includegraphics[width=2.7cm,height=3.5cm]{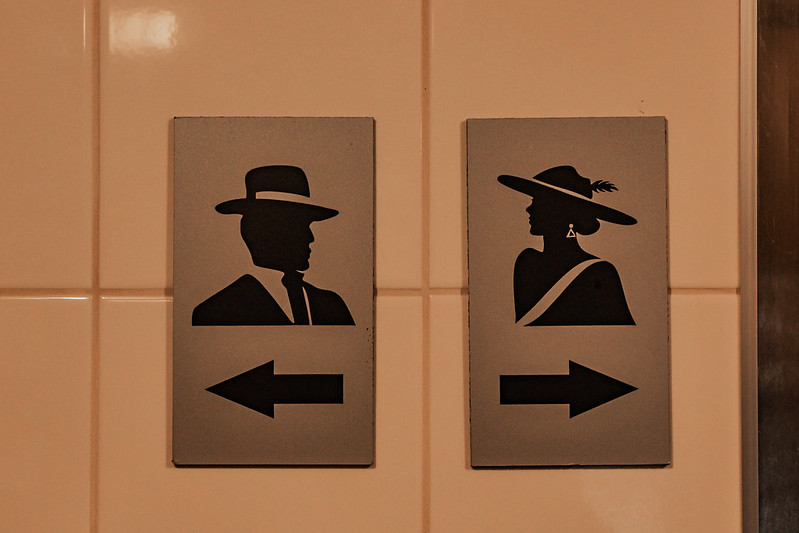}} 
    \subfigure[]{\includegraphics[width=2.7cm,height=3.5cm]{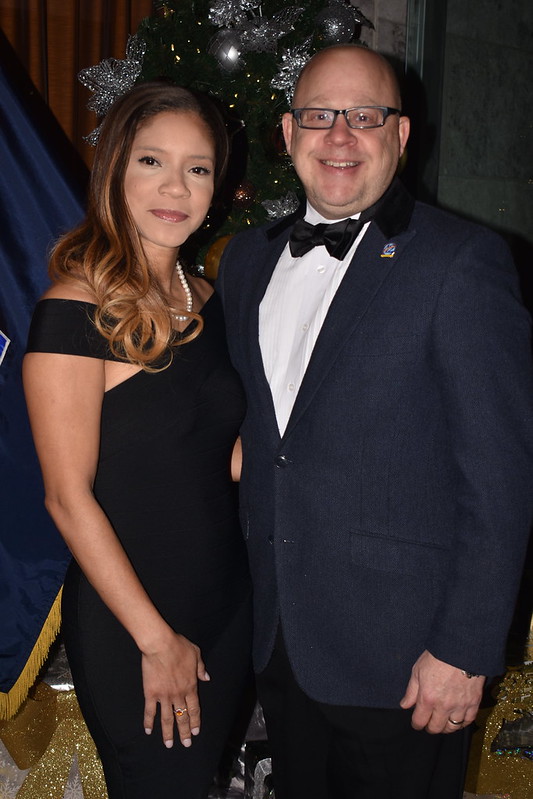}} 
    \subfigure[]{\includegraphics[width=2.7cm,height=3.5cm]{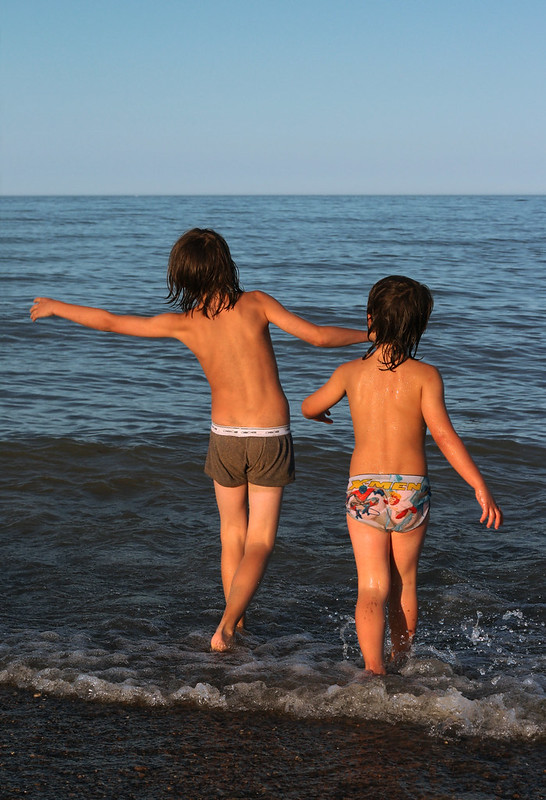}} 
    \caption{Example of miss-classified images from the PrivacyAlert2~\cite{privacyalert} dataset.
    We list below the miss-classifying models (GPA and/or GIP$^{tr}$) and their incorrect prediction.
    (a) GIP$^{tr}$: {public}, (b) GIP$^{tr}$: {private}, (c) GIP$^{tr}$: {public}, (d) GIP$^{tr}$: {private}, (e) GIP$^{tr}$, GPA: {private}, (f) GIP$^{tr}$, GPA: public (see main text for details).}

    \label{fig:misclassif}
\end{figure}

Figure~\ref{fig:misclassif} shows the misclassification results for GIP$^{tr}$ and GPA on the PrivacyAlert2 dataset.
Figure~\ref{fig:misclassif} (a), (b), (c) and (d) are GIP$^{tr}$ classification failures that are correctly predicted by GPA. Figure~\ref{fig:misclassif} (a) and (c) both illustrate the same object (\emph{person}) with the same cardinality (one) and are labelled private. Although prior knowledge in GIP$^{tr}$ strongly associates the presence of a \emph{person} with the private class, GIP$^{tr}$ fails to predict the correct label in both cases. Figure~\ref{fig:misclassif} (b) and (d) are both public images. We hypothesise that GIP$^{tr}$ failed in (d) due to the presence of the \emph{person} icons. Figure~\ref{fig:misclassif} (e) and (f) show examples of images where both methods have failed to provide an accurate prediction on the privacy class of the image. Both images (e) and (f) depict the same object (\emph{person}) with the same cardinality (two); however, (e) is labelled public, whereas (f) is private. Scene information does not provide discriminatory information in this context, underlining the limitations in capturing the privacy risks in both images.

\section{Conclusion}

In this paper, we determine the most relevant image information to be used to improve image privacy classification.
Our proposed graph model, GPA, uses the cardinality of objects and the scene type depicted in the image as core information. 
We also use prior knowledge on object co-occurrences to initialise the graph. 
Future work will consider local information on semantically important regions of an image and further reduce the size of the model.

\section*{Acknowledgement}
This work was supported in part by the CHIST-ERA programme through the project GraphNEx, under UK EPSRC grant EP/V062107/1. 
The authors would also like to thank Vandana Rajan for her help in reproducing the code for GIP. 

\bibliographystyle{IEEEtran}
\bibliography{IEEEabrv, egbib}

\begin{thebibliography}{10}
\providecommand{\url}[1]{#1}
\csname url@samestyle\endcsname
\providecommand{\newblock}{\relax}
\providecommand{\bibinfo}[2]{#2}
\providecommand{\BIBentrySTDinterwordspacing}{\spaceskip=0pt\relax}
\providecommand{\BIBentryALTinterwordstretchfactor}{4}
\providecommand{\BIBentryALTinterwordspacing}{\spaceskip=\fontdimen2\font plus
\BIBentryALTinterwordstretchfactor\fontdimen3\font minus
  \fontdimen4\font\relax}
\providecommand{\BIBforeignlanguage}[2]{{%
\expandafter\ifx\csname l@#1\endcsname\relax
\typeout{** WARNING: IEEEtran.bst: No hyphenation pattern has been}%
\typeout{** loaded for the language `#1'. Using the pattern for}%
\typeout{** the default language instead.}%
\else
\language=\csname l@#1\endcsname
\fi
#2}}
\providecommand{\BIBdecl}{\relax}
\BIBdecl

\bibitem{DRS}
L.~Ferrarello, A.~Cavallaro, R.~Fiadeiro, and R.~Mazzon, ``Reframing the
  {Narrative of Privacy Through System-Thinking Design},'' in \emph{Proc.
  {DRS}}, 2022.

\bibitem{awareness1}
V.~K. Tuunainen, O.~Pitk{\"a}nen, and M.~Hovi, ``Users' {Awareness of Privacy
  on Online Social Networking Sites - Case Facebook},'' in \emph{Bled
  eConference}, 2009.

\bibitem{awareness2}
Y.~Wang, G.~Norcie, S.~Komanduri, A.~Acquisti, P.~G. Leon, and L.~F. Cranor,
  ``"{I Regretted the Minute I Pressed Share": A Qualitative Study of Regrets
  on Facebook},'' in \emph{Proc. of the Seventh Symposium on Usable Privacy and
  Security}, 2011.

\bibitem{picalert2012}
S.~Zerr, S.~Siersdorfer, and J.~Hare, ``{PicAlert! A System for Privacy-Aware
  Image Classification and Retrieval},'' in \emph{Proc. CIKM}, 2012, p.
  2710–2712.

\bibitem{sift}
D.~Lowe, ``Object {Recognition from Local Scale-invariant Features},'' in
  \emph{Proc. {ICCV}}, 1999, pp. 1150--1157.

\bibitem{atongecaragea2019}
A.~Tonge and C.~Caragea, ``{Dynamic Deep Multi-Modal Fusion for Image Privacy
  Prediction},'' in \emph{Proc. {WWW}}, 2019, p. 1829–1840.

\bibitem{atonge_caragea2018}
A.~Tonge, C.~Caragea, and A.~Squicciarini, ``{Uncovering Scene Context for
  Predicting Privacy of Online Shared Images},'' in \emph{Proc. {AAAI}},
  vol.~32, no.~1, 2018.

\bibitem{privacyalert}
C.~Zhao, J.~Mangat, S.~Koujalgi, A.~Squicciarini, and C.~Caragea,
  ``Privacy{Alert: A Dataset for Image Privacy Prediction},'' in \emph{Proc.
  {AAAI}}, vol.~16, no.~1, 2022, pp. 1352--1361.

\bibitem{vgg-16}
S.~Ren, K.~He, R.~Girshick, and J.~Sun, ``{Faster R-CNN: Towards Real-Time
  Object Detection with Region Proposal Networks},'' in \emph{Proc. NIPS},
  vol.~28, 2015.

\bibitem{Xioufis2016PersonalizedPI}
E.~S. Xioufis, S.~Papadopoulos, A.~D. Popescu, and Y.~Kompatsiaris,
  ``Personalized {Privacy-aware Image Classification},'' in \emph{Proc.
  {ICMR}}, 2016.

\bibitem{PCNH}
L.~Tran, D.~Kong, H.~Jin, and J.~Liu, ``{Privacy-CNH: A Framework to Detect
  Photo Privacy with Convolutional Neural Network using Hierarchical
  Features},'' in \emph{Proc. {AAAI}}, 2016.

\bibitem{cooccurence}
J.~Yu, B.~Zhang, Z.~Kuang, D.~Lin, and J.~Fan, ``{iPrivacy: Image Privacy
  Protection by Identifying Sensitive Objects via Deep Multi-Task Learning},''
  \emph{IEEE Transactions on Information Forensics and Security}, vol.~12,
  no.~5, pp. 1005--1016, 2017.

\bibitem{GIP}
G.~Yang, J.~Cao, Z.~Chen, J.~Guo, and J.~Li, ``{Graph-based Neural Networks for
  Explainable Image Privacy Inference},'' \emph{Pattern Recognition}, vol. 105,
  2020.

\bibitem{vispr}
T.~Orekondy, B.~Schiele, and M.~Fritz, ``{Towards a Visual Privacy Advisor:
  Understanding and Predicting Privacy Risks in Images},'' in \emph{Proc.
  ICCV}, 2017, pp. 3706--3715.

\bibitem{euprivacy}
{The European Parliament and Council}, ``{Data Protection Directive
  95/46/EC},'' \emph{Official Journal of the European Union L281}, Oct. 1995.

\bibitem{usprivacy}
{The 93rd United States Congress}, ``{Privacy Act of 1974},'' \emph{Title 5 of
  United States Code (5 U.S.C. 552a)}, Dec. 1974.

\bibitem{socialrule1}
R.~Gross and A.~Acquisti, ``Information {Revelation and Privacy in Online
  Social Networks},'' in \emph{Proc. {WPES}}, 2005, p. 71–80.

\bibitem{drag}
G.~Yang, J.~Cao, Q.~Sheng, P.~Qi, X.~Li, and J.~Li, ``{DRAG: Dynamic
  Region-Aware GCN for Privacy-Leaking Image Detection},'' in \emph{Proc.
  {AAAI}}, 2022.

\bibitem{GSNN}
K.~Marino, R.~Salakhutdinov, and A.~K. Gupta, ``The {More You Know: Using
  Knowledge Graphs for Image Classification},'' in \emph{Proc. {CVPR}}, 2017,
  pp. 20--28.

\bibitem{gnn}
F.~Scarselli, M.~Gori, A.~C. Tsoi, M.~Hagenbuchner, and G.~Monfardini, ``The
  {Graph Neural Network Model},'' \emph{IEEE Transactions on Neural Networks},
  vol.~20, no.~1, pp. 61--80, 2009.

\bibitem{ggnn}
Y.~Li, D.~Tarlow, M.~Brockschmidt, and R.~S. Zemel, ``{Gated Graph Sequence
  Neural Networks},'' in \emph{Proc. {ICLR}}, 2016.

\bibitem{gru}
K.~Cho, B.~van Merrienboer, D.~Bahdanau, and Y.~Bengio, ``On the properties of
  neural machine translation: Encoder–decoder approaches,'' in \emph{Proc.
  {EMNLP}}, 2014.

\bibitem{GRM}
Z.~Wang, T.~Chen, J.~Ren, W.~Yu, H.~Cheng, and L.~Lin, ``{Deep Reasoning with
  Knowledge Graph for Social Relationship Understanding},'' in \emph{Proc.
  IJCAI}, 2018, p. 1021–1028.

\bibitem{gat}
P.~Veličković, G.~Cucurull, A.~Casanova, A.~Romero, P.~Liò, and Y.~Bengio,
  ``{Graph Attention Networks},'' in \emph{Proc. ICLR}, 2018.

\bibitem{gcn}
T.~N. Kipf and M.~Welling, ``Semi-supervised classification with graph
  convolutional networks,'' in \emph{Proc. {ICLR}}, 2017.

\bibitem{sceneplaces}
B.~Zhou, A.~Lapedriza, A.~Torralba, and A.~Oliva, ``{Places: An Image Database
  for Deep Scene Understanding},'' \emph{Journal of Vision}, vol.~17, no.~10,
  2017.

\bibitem{long-tail}
X.~Zhu, D.~Anguelov, and D.~Ramanan, ``Capturing {Long-Tail Distributions of
  Object Subcategories},'' in \emph{Proc. {CVPR}}, 2014, pp. 915--922.

\bibitem{adam}
D.~Kingma and J.~Ba, ``{Adam: A Method for Stochastic Optimization},'' in
  \emph{Proc. {ICLR}}, 2015.

\bibitem{pytorch}
A.~Paszke, S.~Gross, F.~Massa, A.~Lerer, J.~Bradbury, G.~Chanan, T.~Killeen,
  Z.~Lin, N.~Gimelshein, L.~Antiga, A.~Desmaison, A.~Kopf, E.~Yang, Z.~DeVito,
  M.~Raison, A.~Tejani, S.~Chilamkurthy, B.~Steiner, L.~Fang, J.~Bai, and
  S.~Chintala, ``{PyTorch: An Imperative Style, High-Performance Deep Learning
  Library},'' in \emph{Proc. {NIPS}}, 2019.

\bibitem{resnet101}
K.~He, X.~Zhang, S.~Ren, and J.~Sun, ``{Deep Residual Learning for Image
  Recognition},'' in \emph{Proc. CVPR}, 2016, pp. 770--778.

\bibitem{imagenet}
O.~Russakovsky, J.~Deng, H.~Su, J.~Krause, S.~Satheesh, S.~Ma, Z.~Huang,
  A.~Karpathy, A.~Khosla, M.~Bernstein, A.~C. Berg, and L.~Fei-Fei, ``{ImageNet
  Large Scale Visual Recognition Challenge},'' \emph{International Journal of
  Computer Vision (IJCV)}, vol. 115, no.~3, pp. 211--252, 2015.

\bibitem{yolov3}
J.~Redmon and A.~Farhadi, ``{YOLOv3: An Incremental Improvement},''
  \emph{CoRR}, vol. abs/1804.02767, 2018.

\bibitem{coco}
T.-Y. Lin, M.~Maire, S.~Belongie, J.~Hays, P.~Perona, D.~Ramanan,
  P.~Doll{\'a}r, and C.~L. Zitnick, ``{Microsoft COCO: Common Objects in
  Context},'' in \emph{Proc. ECCV}, 2014, pp. 740--755.

\end{thebibliography}

\end{document}